\documentclass[runningheads]{llncs}
\usepackage{graphicx}
\usepackage{amssymb}
\usepackage{amsfonts}
\usepackage{amsmath}
\usepackage{booktabs}
\usepackage{multirow}
\usepackage{multicol}
\usepackage{sidecap}
\usepackage{hyperref}
\usepackage{algorithm2e}
\usepackage{multicol}
\usepackage{multirow}
\usepackage{makecell}
\usepackage{diagbox}
\usepackage{caption}
\usepackage[table]{xcolor}
\usepackage{subcaption}
\usepackage{mathtools}
\usepackage{pdflscape}
\usepackage{pgfplots}
\usepackage{dsfont}
\usepackage{adjustbox}
\usepackage{pifont}
\newif\ifblackandwhitecycle

\allowdisplaybreaks
\newcommand{\name}{SGN}
\newcommand{\ie}{\text{i.e.}}
\newcommand{\eg}{\textit{e.g.}}

\usepackage[capitalize]{cleveref}
\crefname{section}{Sec.}{Secs.}
\Crefname{section}{Section}{Sections}
\Crefname{table}{Table}{Tables}
\crefname{table}{Tab.}{Tabs.}
\Crefname{equation}{Equation}{Equations}
\crefname{equation}{Eq.}{Eqs.}
\Crefname{algorithm}{Algorithm}{Algorithms}
\crefname{algorithm}{Alg.}{Algs.}
\UseRawInputEncoding

\pgfplotsset{ 
    compat=newest,
    scaled y ticks=false,
}
\definecolor{forestgreen}{RGB}{34,139,34}
\definecolor{mypurple}{RGB}{136,43,141}
\definecolor{myorange}{RGB}{240,152,55}
\renewcommand{\paragraph}{\noindent\textbf}

\begin{document}

\title{Spatial Transcriptomics Analysis of Zero-shot Gene Expression Prediction}

\titlerunning{Spatial Transcriptomics Analysis of Zero-shot Gene Expression Prediction}

\author{Yan Yang\inst{1} \and
Md Zakir Hossain\inst{1,2} \and\\
Xuesong Li \inst{1} \and  Shafin Rahman\inst{3} \and Eric Stone\inst{1}}
%
% \authorrunning{Yan Yang et al.}
\institute{Biological Data Science Institute, The Australian National University \and
Optus Centre for AI, Curtin University \and
Department of Electrical and Computer Engineering, North South University
\email{u6169130@anu.edu.au}
}

\authorrunning{Yan Yang et al.}
\maketitle              % typeset the header of the contribution

\begin{abstract}
Spatial transcriptomics (ST) captures gene expression within distinct regions (\ie, windows) of a tissue slide. Traditional supervised learning frameworks applied to model ST are constrained to predicting expression from slide image windows for gene types seen during training, failing to generalize to unseen gene types. To overcome this limitation, we propose a semantic guided network (\name), a pioneering zero-shot framework for predicting gene expression from slide image windows. 
Considering a gene type can be described by functionality and phenotype, we dynamically embed a gene type to a vector per its functionality and phenotype, and employ this vector to project slide image windows to gene expression in feature space, unleashing zero-shot expression prediction for unseen gene types. The gene type functionality and phenotype are queried with a carefully designed prompt from a pre-trained large language model (LLM). On standard benchmark datasets, we demonstrate competitive zero-shot performance compared to past state-of-the-art supervised learning approaches.
\end{abstract}

\section{Introduction}
Spatial transcriptomics (ST) facilitates the exploration and diagnosis of diseases, providing gene expression for fine-grained regions, referred to as windows, in tissue slides. However, acquiring gene expression data for tissue slide windows involves resource-intensive experiments utilizing specialized equipment, typically operated by human experts \cite{st}. This inevitably presents a challenge for collecting datasets for training end-to-end neural networks to predict gene expression from windows of easily obtainable tissue slide images. 
Furthermore, after deploying the trained network, a new demand may arise to predict the expression of gene types that are not used/seen in network training, \ie, unseen gene types, necessitating a revisit of the data collection process for network re-training \cite{UnifiedZSL,SurveyZSL,SurveyZSL3}. Therefore, to address these challenges, this paper pioneerly studies zero-shot gene expression prediction from tissue slide image windows. Our method not only enhances the efficiency and effectiveness of gene expression prediction but also accommodates the prediction of unseen gene types.

\begin{figure}[!t]
    \centering
    \includegraphics{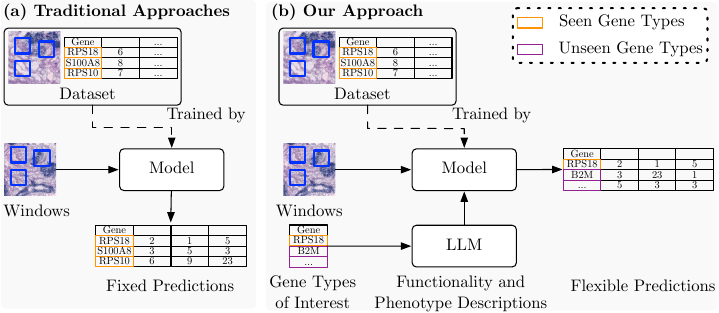}\caption{\small 
    Overview of fields. Training with a dataset of \textcolor{myorange}{seen} gene types, (a) traditional approaches predict the expression of fixed gene types (\ie, \textcolor{myorange}{seen} gene types) for \textcolor{blue}{windows} of a slide image; (b) by using a large language model (LLM) to describe functionality and phenotype of gene types of interest, 
    we flexibly predict expression of \textcolor{myorange}{seen} and \textcolor{mypurple}{unseen} gene types, \ie, zero-shot learning. 
    }
    \label{fig:intro}
\end{figure}

Recently, the computer vision community has been studying gene expression prediction of tissue slide image windows from two perspectives: individual and joint gene expression prediction. In individual prediction approaches, such as those proposed by \cite{stnet,nsl,egn,hsanet,cfanet}, networks are trained to predict gene expression of each window independently. While these approaches have demonstrated promising results, they neglect spatially nearby windows in the slide image often share similar gene expression that could mutually benefit each other in the prediction task.

Considering this insight, the core of joint gene expression prediction, as outlined in \cite{eggn2023,SEPAL}, is to embed each window of a tissue slide image into features, connect windows as a graph, apply graph convolutions networks \cite{gcns,sage} for establishing dependency among windows for refining window features, and predict gene expression of each window from the refined window features. The graph is constructed by treating each window as a node and connecting edges among spatially nearby window nodes. However, the similar windows within a slide image could also mutually benefit each other for each gene expression prediction, which is underexplored in past works.

Nevertheless, existing individual and joint gene expression prediction approaches focus on traditional supervised learning, thereby restricting expression prediction to gene types seen during training. For example, 250 common gene types with the highest expression are intentionally selected for training on the STNet dataset \cite{stnet,nsl,egn,hsanet,cfanet,eggn2023,SEPAL}, and the past methods cannot make predictions on remaining rare gene types \cite{stnet}.
This paper takes a pioneering step forward by studying zero-shot gene expression prediction of windows in a tissue slide image, extending the prediction ability to unseen gene types by presenting our semantic guided network (\name). Our key idea is to describe a gene type by its functionality and phenotype and use the description to project tissue slide windows to gene expression in feature space. 

Formally, our \name~implements zero-shot gene expression prediction in three stages. Firstly, we extract a feature vector for each window using a pre-trained network. Inspired by  \cite{eggn2023}, in a slide image, we treat each window as a node, and construct a homogeneous graph that connects nearby windows and windows with similar extracted features. A graph convolution network \cite{sage} is then applied to refine the features of each window, benefiting from spatially nearby and feature-similar windows. Concurrently, to obtain the functionality and phenotype descriptions of the gene type of interest, we design a prompt to leverage a pre-trained large language model (LLM) for querying the description. As a general-purpose LLM potentially lacks domain-specific gene type knowledge, when internet access is available, we automatically scrap references related to the gene type to supplement the knowledge of the LLM for describing gene type functionality and phenotype. We then embed the gene type description into a projection vector by using a neural network. Finally, we perform a dot product between window features and gene type projection vector to derive the gene expression. Experimentally, our method achieves competitive performance with the state-of-the-art traditionally supervised approach on standard benchmark datasets, evaluating our method on unseen gene types.

\section{Method}

\begin{figure}[!t]
    \centering
    \includegraphics{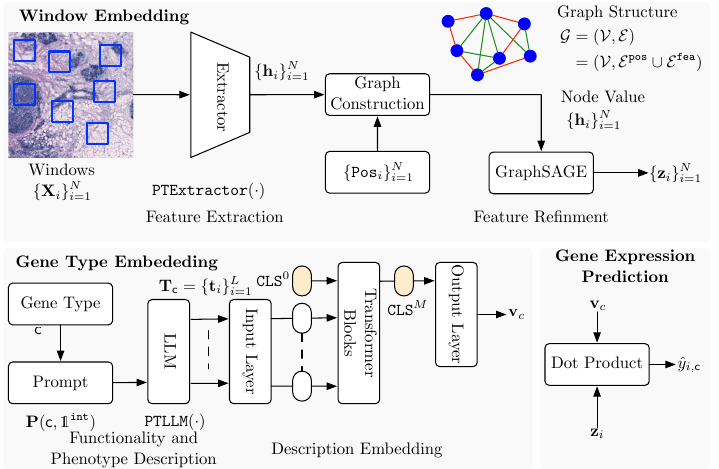}
    \caption{\small Our framework. We have three stages:
    i) window embedding, extracting and refining features from each \textcolor{blue}{window} by using an extractor and a GraphSAGE network that explores relations of window \textcolor{red}{spatial positions} and \textcolor{forestgreen}{feature similarities}; ii) gene type embedding, querying gene type functionality and phenotype from an LLM, and embedding the description; iii) gene type prediction, performing dot product between window embedding and gene type embedding to compute gene expression. 
    }
    \label{fig:method}
\end{figure}

\paragraph{Overview.} We distinguish gene types into seen gene types $\mathcal{C}^{\mathsf{s}}$ and unseen gene types $\mathcal{C}^{\mathsf{u}}$, ensuring $\mathcal{C}^{\mathsf{s}} \cap \mathcal{C}^{\mathsf{u}} = \emptyset$. Given a slide image containing $N$ windows $\{\mathbf{X}_{i}\}_{i=1}^{N}$ and the gene type $\mathsf{c} \in \mathcal{C}^{\mathsf{s}} \cup \mathcal{C}^{\mathsf{u}}$ of interest, our goal is to predict the expression $\{\hat{y}_{i, \mathsf{c}}\}_{i=1}^{N}$ of gene type $\mathsf{c}$ for all windows $\{\mathbf{X}_{i}\}_{i=1}^{N}$ in the slide image, where $\mathbf{X}_{i} \in \mathbb{R}^{H \times W \times 3}$, $\hat{y}_{i, \mathsf{c}} \in \mathbb{R}$, and $H$ and $W$ are height and width of the window $\mathbf{X}_{i}$, respectively. This zero-shot learning framework is trained by using ground truth expression $\{y_{i,\mathsf{c}}\}_{i=1}^{N}$ of gene type $\mathsf{c} \in \mathcal{C}^{\mathsf{s}}$ to supervise the predicted expression $\{\hat{y}_{i,\mathsf{c}}\}_{i=1}^{N}$, and has three steps: i) window embedding, composing of feature extraction and refinement to obtain $D$-dimensional feature vectors $\{\mathbf{z}_{i}\}_{i=1}^{N}$, \ie, $\mathbf{z}_{i} \in \mathbb{R}^{1 \times D}$; ii) gene type embedding, obtaining functionality and phenotype descriptions $\mathbf{T}_{c}$ of gene type $\mathsf{c}$ by using a pre-trained LLM, and deriving a projection vector $\mathbf{v}_{\mathsf{c}} \in \mathbb{R}^{1 \times D}$ for gene type $\mathsf{c}$ per the descirbe $\mathbf{T}_{c}$;
iii) gene expression prediction, computing gene expression by applying a dot product $\{\hat{y}_{i,\mathsf{c}}\}_{i=1}^{N} = \{\mathbf{z}_{i} \cdot {\mathbf{v}_{\mathsf{c}}}^{\top}\}_{i=1}^{N}$. 
In testing, we evaluate our method by repeating the above three steps to predict the expression of the unseen gene type $c \in \mathcal{C}^{\mathsf{u}}$ on the slide image.  
In the remaining of the paper, for brevity, we denote a matrix with shape $1 \times 1$ as a scalar, \eg, $\mathbf{z}_{i} \cdot {\mathbf{v}_{\mathsf{c}}}^{\top} \in \mathbb{R}$. We show the overall framework in \cref{fig:method}. 

\subsection{Window Embedding}
We use graph-based settings to embed the windows $\{\mathbf{X}_{i}\}_{i=1}^{N}$ into $\{\mathbf{z}_{i}\}_{i=1}^{N}$ through feature extractions and refinements, providing discriminative features for our zero-shot gene expression prediction task.

\paragraph{Feature Extraction.} We use a pre-trained network \cite{resnet}, $\texttt{PTExtractor}(\cdot)$, to extract window features $\{\mathbf{h}_{i}\}_{i=1}^{N} = \{\texttt{PTExtractor}(\mathbf{X}_{i})\}_{i=1}^{N}$, where  $\mathbf{h}_{i} \in \mathbb{R}^{1 \times D^{\mathbf{e}}}$ and $D^{\mathsf{e}}$ is the feature dimension of the pre-trained network $\texttt{PTExtractor}(\cdot)$. However, the window features $\{\mathbf{h}_{i}\}_{i=1}^{N}$ are independently and locally extracted. They are short of context/global information of the slide image which has been proven to be beneficial for predicting gene expression of each window \cite{eggn2023}, motivating us to perform feature refinement for each window in the next section.

\paragraph{Feature Refinement.} We define a graph $\mathcal{G} = (\mathcal{V}, \mathcal{E})$ with a node set $\mathcal{V}$ and an edge set $\mathcal{E}$ for windows in the slide image, and use graph convolution network to refine window features $\{\mathbf{h}_{i}\}_{i=1}^{N}$ base on the graph structure $\mathcal{G}$.  

To construct $\mathcal{G} = (\mathcal{V}, \mathcal{E})$, we set each window $\mathbf{X}_{i}$ as a node to form the node set $\mathcal{V} = \{v_{i}\}_{i=1}^{N}$, and consider two edge types, $\mathcal{E} = \mathcal{E}^{\texttt{pos}} \cup \mathcal{E}^{\texttt{fea}}$ to explore relations of spatial position and feature similarity among windows, \ie, the two edge type gives context/global relations among the graph. With $\texttt{Pos}_{i} \in \mathbb{R}^{2}$ describing spatial position of each window $\mathbf{X}_{i}$ in the slide image, we have 
\begin{align}
    \mathcal{E}^{\texttt{pos}} & = \{e_{ij} \mid v_{i}, v_{j} \in \mathcal{V} \times \mathcal{V} \land \phi(\texttt{Pos}_{i}, \texttt{Pos}_{j}, \{\texttt{Pos}_{k}\}_{k=1}^{N} ) \}  \ ,\\
    \mathcal{E}^{\texttt{fea}} & = \{e_{ij} \mid v_{i}, v_{j} \in \mathcal{V} \times \mathcal{V} \land \phi(\mathbf{h}_{i}, \mathbf{h}_{j}, \{\mathbf{h}_{k}\}_{k=1}^{N} ) \} \ .
\end{align}
$\xi$ is a k-nearest neighbors (k-NN) function, determining  nearest neighbors, \eg, $\phi(\texttt{Pos}_{i}, \texttt{Pos}_{j}, \{\texttt{Pos}_{k}\}_{k=1}^{N} )$ computes if $\texttt{Pos}_{i}$ is one of the nearest neighbors of $\texttt{Pos}_{j}$ among $\{\texttt{Pos}_{k}\}_{k=1}^{N}$.

With the graph $\mathcal{G} = (\mathcal{V}, \mathcal{E})$, we refine the window features $\{\mathbf{h}_{i}\}_{i=1}^{N}$ into $\{\mathbf{z}_{i}\}_{i=1}^{N}$ with a GraphSAGE network \cite{sage} along the two edges $\mathcal{E}^{\tt{pos}} \cup \mathcal{E}^{\tt{fea}}$. Mathematically, the refined features $\{\mathbf{z}_{i}\}_{i=1}^{N}$ is defined by
\begin{align}
    \mathbf{z}_{i}  =  \left[\mathbf{h}_{i} \mathbin \Big\Vert \frac{1}{\lvert \mathcal{N}^{\texttt{pos}}_{i} \rvert} \sum_{j \in \mathcal{N}^{\texttt{pos}}_{i}} \mathbf{h}_{j} \mathbin \Big\Vert \frac{1}{\lvert \mathcal{N}^{\texttt{fea}}_{i} \rvert} \sum_{k \in \mathcal{N}^{\texttt{feat}}_{i}} \mathbf{h}_{k}  \right] ~ \mathbf{W}^{\mathbf{z}} \ , \label{eq:sage} 
\end{align}
where $\mathbf{W}^{\mathbf{z}} \in \mathbb{R}^{(3 \times \mathbf{D}^{\mathbf{e}}) \times D}$ is a linear weight matrix, [$\cdot \mathbin\Vert \cdot \mathbin\Vert \cdot$] is a concatenation operator,  and $\mathcal{N}^{\texttt{pos}}_{i} = \{j \mid e_{jk} \in \mathcal{E}^{\texttt{pos}} \land k = i \}$ and $\mathcal{N}^{\texttt{fea}}_{i} = \{k \mid e_{kj} \in \mathcal{E}^{\texttt{fea}} \land j = i \}$ 
gives indexes of nodes connected to $v_{i}$ along $\mathcal{E}^{\tt{pos}}$ and $\mathcal{E}^{\tt{fea}}$, respectively. The refined features $\{\mathsf{z}_{i}\}_{i=1}^{N}$ are finally used to perform our zero-shot gene expression prediction in \cref{sec:predcit}.

\subsection{Gene Type Embedding}
We generate functionality and phenotype description for a gene type $\mathsf{c}$, and dynamically embed the description into a vector $\mathbf{v}_{\mathsf{c}}$ that can project refined window features $\{\mathsf{z}_{i}\}_{i=1}^{N}$ into the gene type expression. Refer to our supplementary materials for generated descriptions.

\paragraph{Functionality and Phenotype Description.} To generate descriptions of functionality and phenotype for a gene type $\mathsf{c}$, we leverage a pre-trained LLM, and design a prompt $\mathbf{P}(\mathsf{c}, \mathds{1}^{\texttt{int}})$ that is adjusted by internet access availability $\mathds{1}^{\texttt{int}}$. If the internet access is absent, \ie, $\mathds{1}^{\texttt{int}}=\texttt{false}$, the prompt
$\mathbf{P}(\mathsf{c}, \mathds{1}^{\texttt{int}})$ directly query the functionality and phenotype description from the LLM. Conversely, with the internet access, \ie, $\mathds{1}^{\texttt{int}}=\texttt{true}$, we supplement the knowledge base of the LLM by providing a domain-specific gene type reference, conditionally prompting the LLM to generate the functionality and phenotype description. The generated description $\mathbf{T}_{\mathsf{c}} $ are obtained by 
\begin{align}
    \mathbf{T}_{\mathsf{c}} = \texttt{PTLLM}(\mathbf{P}(\mathsf{c}, \mathds{1}^{\texttt{int}})) \ , \quad \mathbf{T} \in \mathbb{R}^{L \times D^{\mathbf{T}}} \ . 
\end{align}
Here, $\texttt{PTLLM}(\cdot)$ is a pre-trained LLM model \cite{neural-chat}, $L$ is the length of the description, and $D^{\mathbf{T}}$ is the feature dimension of $\texttt{LLM}(\cdot)$. We preserve the feature representation capability of the $\texttt{LLM}(\cdot)$ by discarding its classification layer, using the final embedding layer output as our description $\mathbf{T}_{\mathsf{c}}$.

\paragraph{Description Embedding.} We embed $\mathbf{T}_{\mathsf{c}}$ into a vector $\mathbf{v}_{\mathsf{c}}$ by using a transformer \cite{vit}, aligning them to a joint feature space of the refined window features $\{\mathbf{z}_{i}\}$, and summarising information beneficial to the expression prediction of gene type $\mathsf{c}$. The transformer has an input layer, a list of transformer blocks, and an output layer in order. The computation is described as follows. We have a input layer that project $\mathbf{T}_{\mathsf{c}}$ to a $D^{\mathbf{E}}$-dimensional matrix by using a weight matrix $\mathbf{W}^{\mathbf{T}_{c}} \in \mathbf{R}^{D^{\mathbf{T}} \times D^{\mathbf{E}}}$, and append a [\texttt{CLS}$^{0}$] token to the projected matrix, obtaining $\mathbf{E}^{0}_{\mathsf{c}}$, 
\begin{align}
    \mathbf{E}^{0}_{\mathsf{c}} = \{\texttt{CLS}^{0}, \mathbf{t}_{1} \mathbf{W}^{\mathbf{T}_{c}}, \mathbf{t}_{2} \mathbf{W}^{\mathbf{T}_{c}}, \cdots, \mathbf{t}_{L} \mathbf{W}^{\mathbf{T}_{c}} \} \ ,
\end{align}
where $\mathbf{t}_{i}$ is the $i$-th token embedding of $\mathbf{T}_{\mathsf{c}}$, \ie, $\mathbf{T}_{\mathsf{c}} = \{\mathbf{t}_{i}\}_{i=1}^{L}$. Assuming there are $M$ transformer blocks, for $1\leq m \leq M$, we compute embedding $\mathbf{E}^{m}_{\mathsf{c}}$ of the $m$-th block as  
\begin{align}
    \mathbf{E}^{m}_{\mathsf{c}} = \texttt{FFN}^{m}\left(\texttt{Attention}^{m}\left(\mathbf{E}^{m-1}_{\mathsf{c}}, \mathbf{E}^{m-1}_{\mathsf{c}}, \mathbf{E}^{m-1}_{\mathsf{c}} \right)\right) \ .
\end{align}
$\texttt{FFN}^{m}(\cdot)$ and $\texttt{Attention}^{m}(\cdot,\cdot,\cdot)$ are respectively a feedforward layer and an attention layer in the $m$-th block \cite{vit}. In an output layer, we then pop out the [\texttt{CLS}$^{M}$] token from  $\mathbf{E}^{M}_{\mathsf{c}}$, and project it to $\mathbf{v}_{\mathsf{c}}$ by using a weight matrix $\mathbf{W}^{\mathbf{v}} \in \mathbb{R}^{D^{\mathbf{E}} \times D}$.

\subsection{Gene Expression Prediction}
\label{sec:predcit}
With refined window features $\{\mathbf{z}_{i}\}_{i=1}^{N}$ and gene type embedding $\mathbf{v}_{\mathsf{c}}$ in a shared feature space, $\mathbf{v}_{\mathsf{c}}$ is used to project each $\mathbf{z}_{i}$ for performing zero-shot expression prediction of gene type $\mathsf{c}$ as  
\begin{align}
    \{\hat{y}_{i,\mathsf{c}}\}_{i=1}^{N} = \{\mathbf{z}_{i} \cdot {\mathbf{v}_{\mathsf{c}}}^{\top}\}_{i=1}^{N} \ . 
\end{align}

\subsection{Loss}
We optimize our network with a mean square error $\mathcal{L}_{\text{mse}}$ and batch-wise Pearson correlation coefficient (PCC) loss $\mathcal{L}_{\text{pcc}}$. The $\mathcal{L}_{\text{mse}}$ penalize deviations of gene expression predictions $\{\hat{y}_{i,\mathsf{c}}\}_{i=1}^{N}$ from the ground-truth gene expression  $\{y_{i,\mathsf{c}}\}_{i=1}^{N}$. The $\mathcal{L}_{\text{pcc}}$ encourages the correlation between $\{y_{i,\mathsf{c}}\}_{i=1}^{N}$ and $\{y_{i,\mathsf{c}}\}_{i=1}^{N}$. The overall training loss $\mathcal{L}$ is defined as 
\begin{align}
    \mathcal{L} = \mathcal{L}_{\text{mse}} + \mathcal{L}_{\text{pcc}}  \ .
\end{align}

\begin{table}[!t]
    \centering
    \small 
    \caption{\small Quantitative gene expression prediction comparisons with SOTA methods on STNet dataset and 10xProteomic dataset. We present the performance of our method trained by traditional supervised learning as our performance upper bound. For the zero-shot setting, without cherry-picking, we set gene types selected by past works as unseen gene types, and test these gene types.
    }
    \setlength{\tabcolsep}{2.42pt}
    \begin{tabular}{lcccccc}
        \toprule
        Method & Zero-shot & MSE$_{\times10^{2}}$ & MAE$_{\times10^{1}}$ &PCC@F$_{\times10^{1}}$ &PCC@S$_{\times10^{1}}$ &PCC@M$_{\times10^{1}}$  \\
         \midrule
         \rowcolor{orange!10!white}
         \multicolumn{7}{l}{Exiperiments on the STNet dataset.} \\
            STNet \cite{stnet} & \ding{55} &4.52 &1.70 &0.05 &0.92&0.93\\
            NSL \cite{nsl} & \ding{55}  & - & - & -0.71  &0.25  &0.11  \\
           EGN \cite{egn} & \ding{55} & 4.10&1.61&1.51&2.25& 2.02\\
           HSANet \cite{hsanet} & \ding{55} & 4.00 & 1.59 & 1.60 & 2.28 & 2.38 \\
           CFNet \cite{cfanet}  & \ding{55} & 6.30 & 1.66 & 2.12 & 3.06 & 3.00 \\
           EGGN \cite{eggn2023} & \ding{55} &  3.94 & 1.61 & 2.12 & 3.05 & 2.92 \\
           Ours & \ding{55} & 4.38 & 1.72 & 2.00 & 3.03 & 2.83\\
           Ours & \ding{51} & 11.86 & 2.88 & 1.79 & 2.89 & 2.69\\
           \midrule
           \rowcolor{orange!10!white}
           \multicolumn{7}{l}{Exiperiments on the 10xProteomic dataset.} \\
           STNet \cite{stnet} & \ding{55} & 12.40 & 2.64 &1.25 &2.26 &2.15 \\
            NSL \cite{nsl} & \ding{55}  & - & - &-3.73 & 1.84 & 0.25  \\
           EGN \cite{egn} & \ding{55} & 5.49& 1.55& 6.78& 7.21& 7.07 \\
           HSANet \cite{hsanet} & \ding{55} & 4.00 & 1.54 & 6.93 & 7.43 & 7.20  \\
           CFNet \cite{cfanet}  & \ding{55} & 4.00 & 1.49 & 8.00 & 8.16 & 8.02\\
           EGGN \cite{eggn2023} & \ding{55} & 3.52 & 1.31 & 7.06 & 7.60 & 7.44 \\
           Ours & \ding{55} & 4.27 & 1.67 & 8.22 & 8.38 & 8.15\\
           Ours & \ding{51} & 13.05 & 2.70 & 6.33 & 6.51 & 6.48\\
         \bottomrule         
    \end{tabular}
    \label{tab:eva}
\end{table}

\section{Experiment}

\paragraph{Datasets.} We experiment with the STNet dataset \cite{stnet} and 10xProteomic datasets\footnote{\url{https://www.10xgenomics.com/resources/datasets}}. The STNET dataset and 10xProteomic dataset have 30,612 windows from 68 slide images and 24,263 windows from 6 slide images, respectively. We follow the dataset pre-processing and cross-fold validation settings of \cite{egn,eggn2023}. Past works select 250 gene types with the largest mean across the dataset as prediction targets. To compare with the past gene expression prediction works, we use their unselected gene types in training as seen gene types and their selected gene types in testing as unseen gene types. 

\paragraph{Evaluation Metrics.} Our method is evaluated with mean squared error (MSE), mean absolute error (MAE), first quartile of PCC (PCC@F), median of PCC (PCC@S), and  mean of PCC (PCC@M). 

\paragraph{Implementation Details.}  We implement \name~by using the \textit{PyTorch Geometric} \cite{pytorch,pyg} frameworks. We train \name~respectively for $100$ epochs and $300$ epochs on the STNet dataset and 10xProteomic dataset with batch size $1$, where a slide image contains up to thousands of windows in the two datasets. We use the learning rate $5 \times 10^{-4}$ and weight decay $1 \times 10^{-4}$. Follow \cite{eggn2023}, we use a four-layer GraphSAGE  with hidden dimensions 512. For the gene type embedding, a two-layer ViT with hidden dimension 256 is used.

\subsection{Experimental Result}
We compare with state-of-the-art methods on the STNet dataset and 10xProteomic dataset in \cref{tab:eva}. Among all the methods, we are the only method that predicts gene expression in a zero-shot manner. Though our method performs poorly on absolute gene expression prediction evaluation metrics, MSE and MAE, our method successfully captures the relative variations of gene expression across different windows, \ie, PCC@F, PCC@S, and PCC@M. For example, on the STNet dataset, our method finds 0.269 PCC@M that is slightly 0.031 PCC@M lower than the state-of-the-art method, CFNet (3.00 PCC@M). As demonstrated by \cite{egn,eggn2023}, capturing relative variations of gene expression are most important in our task, and our method shows competitive zero-shot gene expression prediction performance against the state-of-the-art traditional supervised learning approaches in these metrics. This validates the performance of our zero-shot gene expression prediction framework.

\begin{table}[!t]
    \centering
    \small 
    \caption{\small Ablation of our model components. We use `FI' and `PT' for shorts of functionality and phenotype.}
    \label{tab:ablation}
    \setlength{\tabcolsep}{10.2pt}
    \begin{tabular}{cccccccc}
    \toprule
     \multicolumn{2}{c}{GraphSAGE} && \multicolumn{2}{c}{LLM} &  \multirow{2}{*}{MSE$_{\times10^{2}}$} & \multirow{2}{*}{MAE$_{\times10^{1}}$} & \multirow{2}{*}{PCC@M$_{\times10^{1}}$} \\
     \cmidrule{1-2} \cmidrule{4-5}
      $\mathcal{E}^{\texttt{pos}}$ & $\mathcal{E}^{\texttt{fea}}$ && FI & PT \\
     \midrule
      \ding{55} & \ding{55} && \ding{51} & \ding{51} & 10.70 & 2.71 & 2.58\\
      \ding{51} & \ding{55} && \ding{51} & \ding{51} & 11.59 & 2.85 &  2.65\\
      \ding{55} & \ding{51} && \ding{51} & \ding{51} & 10.66 & 2.73 & 2.48\\
      \ding{51} & \ding{51} && \ding{55} & \ding{51} & 12.53 & 2.97 & 2.61\\
      \ding{51} & \ding{51} && \ding{51} & \ding{55} & 11.52 & 2.83 & 2.58\\
      \ding{51} & \ding{51} && \ding{51} & \ding{51} & 11.86 & 2.88 & 2.69\\
     \bottomrule
    \end{tabular}
\end{table}

\begin{figure*}[!t]
    \vspace{-1em}
    \centering
    \includegraphics{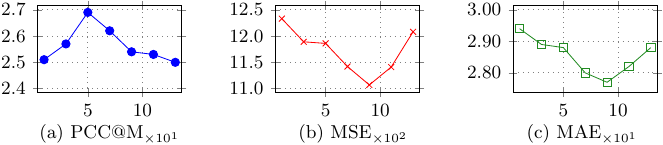}
    \vspace{-.5em}
    \captionof{figure}{Ablation study on the number of neighbors used in constructing $\mathcal{E}^{\texttt{fea}}$. 
    }
    \label{fig:nmk}
\end{figure*}
\begin{figure*}
 \vspace{-1em}
 \centering
 \includegraphics[]{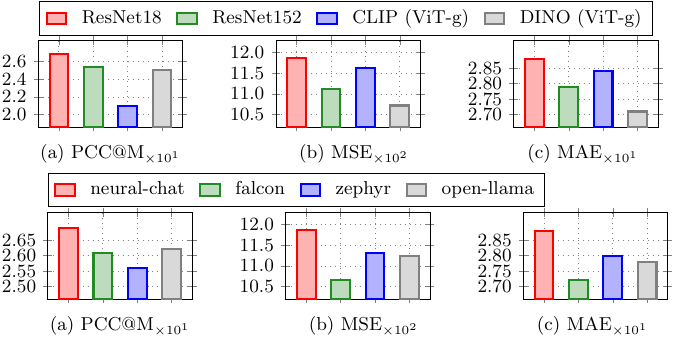}
    \vspace{-.5em}
    \captionof{figure}{Ablation study of the pre-trained feature extractor (a-d) \cite{resnet,clip-vit-g,dino-v2} and LLM (d-f) \cite{neural-chat,falcon,zephyr,open-llama}.
    }
    \label{fig:extractor}
\end{figure*}

\subsection{Ablation}

\paragraph{Model Component.} We ablate the our model components in \cref{tab:ablation}. When disabling the GraphSAGE, a single linear layer is used to unify the feature dimensions. Consistently, our  components improve the prediction performance.   

\paragraph{Number of Neighbors.} We study the number of k-NN edges used for constructing $\mathcal{E}^{\texttt{fea}}$ in \cref{fig:nmk}. Again, we are biased on PCC-based evaluation metrics. Having 5 kNN edges finds the most balanced performance. 

\paragraph{Extractor and LLM.} We ablate the pre-trained feature extractor and LLM in \cref{fig:extractor}. In contrast to recent trends that prefer ViT-g feature extractor for multi-modality feature interaction, with using neural-chat as our pre-trained LLM, our result suggests that using ResNet18 as a feature extractor finds the best performance. 

\paragraph{Reference for LLM.} By default, our model is evaluated by automatically scraping gene type references from the internet. When internet access is disabled, we still have a competitive performance that finds lower performance, \ie, 0.256 PCC@M, 0.124 MAE, and 0.295 MSE.
This shows that our model can robustly predict gene expression.

\section{Conclusion}
This paper studies an \name~framework for zero-shot gene expression prediction of windows in a tissue slide image. Given a gene type of interest, we design a prompt to query the functionality and phenotype of the gene from a pre-trained LLM. The obtained gene type description is then used to project each window to the expression of the gene type in feature space. Finally, we compare our zero-shot gene expression prediction framework with the past state-of-the-art supervised learning approaches, and experimentally demonstrate competitive gene expression prediction performance.

\bibliographystyle{splncs04}
\bibliography{bibliography}

\end{document}